\documentclass[10pt,twocolumn,letterpaper]{article}

\usepackage[]{cvpr}      

\usepackage{graphicx}
\usepackage{amsmath}
\usepackage{amssymb}
\usepackage{booktabs}
\usepackage{makecell}
\usepackage{multirow}

\usepackage[pagebackref,breaklinks,colorlinks]{hyperref}

\usepackage[capitalize]{cleveref}
\crefname{section}{Sec.}{Secs.}
\Crefname{section}{Section}{Sections}
\Crefname{table}{Table}{Tables}
\crefname{table}{Tab.}{Tabs.}

\def\cvprPaperID{8391} 
\def\confName{CVPR}
\def\confYear{2022}

\begin{document}

\title{Attention Concatenation Volume for Accurate and Efficient Stereo Matching}

\author{
Gangwei Xu$^{1,}$\footnotemark[1] ,  Junda Cheng$^{1,}$\footnotemark[1] ,  Peng Guo$^{1}$ ,  Xin Yang$^{1,2,}$\footnotemark[2]\\
$^{1}$School of EIC, Huazhong University of Science \& Technology\\
$^{2}$Wuhan National Laboratory for Optoelectronics\\
{\tt\small \{gwxu, cjd, guopeng, xinyang2014\}@hust.edu.cn}
}


\maketitle

\begin{abstract}
Stereo matching is a fundamental building block for many vision and robotics applications. An informative and concise cost volume representation is vital for stereo matching of high accuracy and efficiency.
In this paper, we present a novel cost volume construction method which generates attention weights from correlation clues to suppress redundant information and enhance matching-related information in the concatenation volume. To generate reliable attention weights, we propose multi-level adaptive patch matching to improve the distinctiveness of the matching cost at different disparities even for textureless regions. The proposed cost volume is named attention concatenation volume (ACV) which can be seamlessly embedded into most stereo matching networks, the resulting networks can use a more lightweight aggregation network and meanwhile achieve higher accuracy, e.g. using only 1/25 parameters of the aggregation network can achieve higher accuracy for GwcNet. Furthermore, we design a highly accurate network (ACVNet) based on our ACV, which
achieves state-of-the-art performance
on several benchmarks. 
The code is available at \textcolor{magenta}{https://github.com/gangweiX/ACVNet}.
\end{abstract}
{
\renewcommand{\thefootnote}{\fnsymbol{footnote}}
\footnotetext[1]{Authors contributed equally.}
\footnotetext[2]{Corresponding author.}}

\section{Introduction}
\label{sec:intro}
Stereo matching which establishes dense correspondences between pixels in a pair of rectified stereo images is a key enabling technique for many applications such as robotics, augmented reality, and autonomous driving. Despite of extensive studies in this field, 
how to concurrently achieve a high inference accuracy and efficiency is critical for practical applications yet remains challenging.

Recently, convolutional neural networks have exhibited great potential in this field~\cite{dispNetC2016large, chang2018pyramid, guo2019group, xu2020aanet}. State-of-the-art CNN stereo models typically consist of four steps, i.e. feature extraction, cost volume construction, 
\begin{figure}[t]
\centering
{\includegraphics[width=1.0\linewidth]{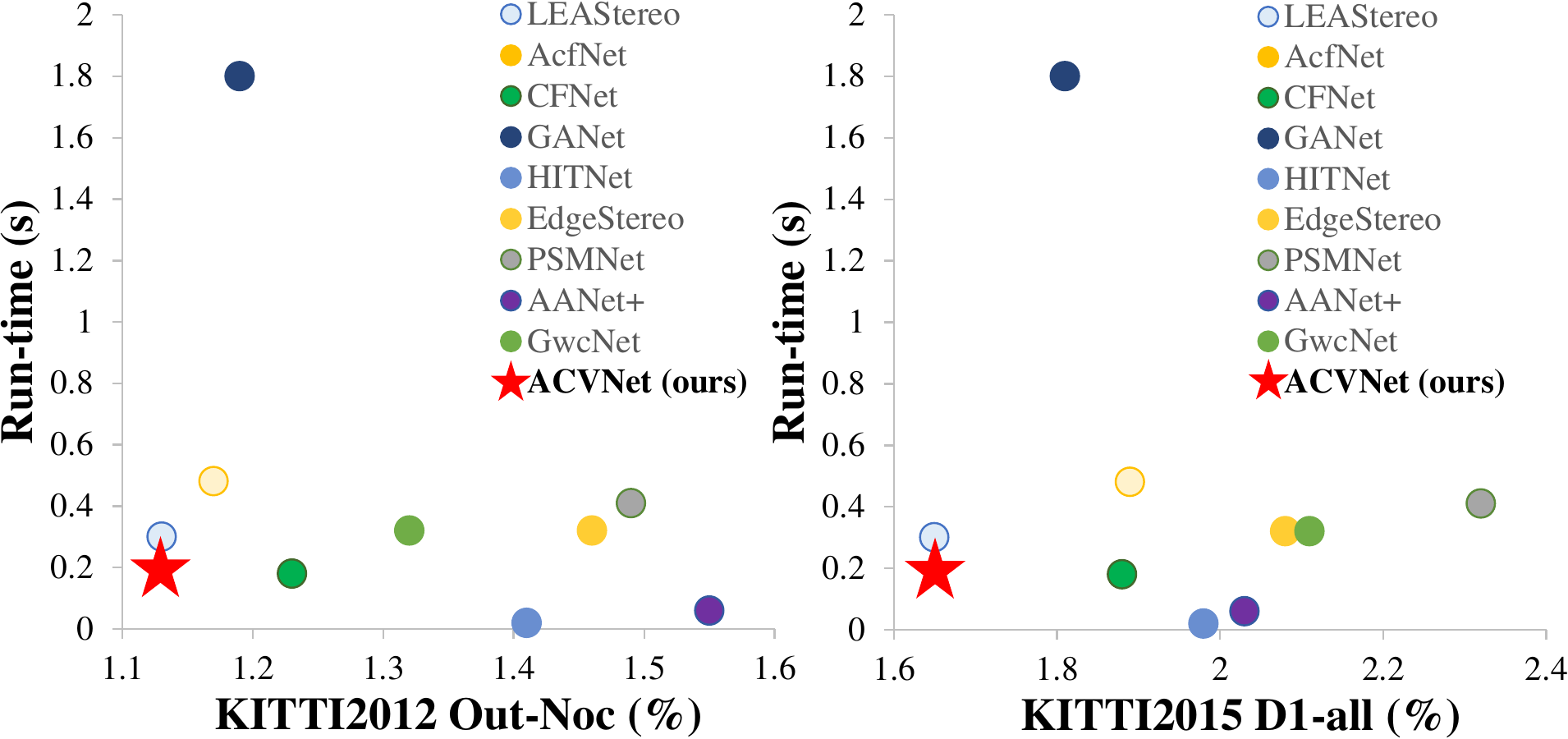}}
\vspace{-15pt}
\caption{
Out-Noc error vs. Run-time on the KITTI 2012 leaderboard and D1-all error vs. Run-time on the KITTI 2015 leaderboard. Our ACVNet, denoted by red stars, achieves competitive performance compared to other 
state-of-the-art stereo models.}\label{fig:ranking}
\vspace{-15pt}
\end{figure}
cost aggregation and disparity regression. Cost volume which provides initial similarity measures for left image pixels and possible corresponding right image pixels is a crucial step of stereo matching. An informative and concise cost volume representation from this step is vital for the final accuracy and computational complexity. Learning-based methods explore different cost volume representations. 
DispNetC~\cite{dispNetC2016large} computes a single-channel full correlation volume between the left and right feature maps. Such full correlation volume provides an efficient way for measuring similarities, but it loses much content information.
GC-Net~\cite{kendall2017end}
constructs a 4D concatenation volume by concatenating left and right feature maps along all disparity levels to provide abundant content information. However, the concatenation volume completely ignores similarity measurements, and thus requires extensive 3D convolutions for cost aggregation to learn similarity measurements from scratch. To tackle the above drawbacks, 
GwcNet~\cite{guo2019group} concatenates the group-wise correlation volume with a compact concatenation volume to encode both matching and content information in the final 4D cost volume. However, the data distribution and characteristics of a correlation volume and a concatenation volume are quite different, i.e. the former represents the similarity measurement obtained through dot product, and the latter is the concatenation of the unary features. Simply concatenating the two volumes and regularizing them via 3D convolutions can hardly exert the advantages of the two volumes to the full. As a result, GwcNet still requires twenty eight 3D convolutions for cost aggregation.

This work aims to explore a more efficient and effective form of cost volume, which can significantly alleviate the burden of cost aggregation
and meanwhile achieve the state-of-the-art accuracy. 
We build our model based on two key observations: first, the concatenation volume contains rich but redundant content information;
second, the correlation volume which measures feature similarities between left and right images can implicitly reflect relationships among neighboring pixels in an image, i.e. nearby pixels which belong to the same class tend to have close similarities. This suggests that utilizing the correlation volume which encodes pixel relationship prior can facilitate a concatenation volume to significantly suppress its redundant information and meanwhile maintain sufficient information for matching in the concatenation volume.

\begin{figure*}
\centering
\includegraphics[width=1.0\textwidth]{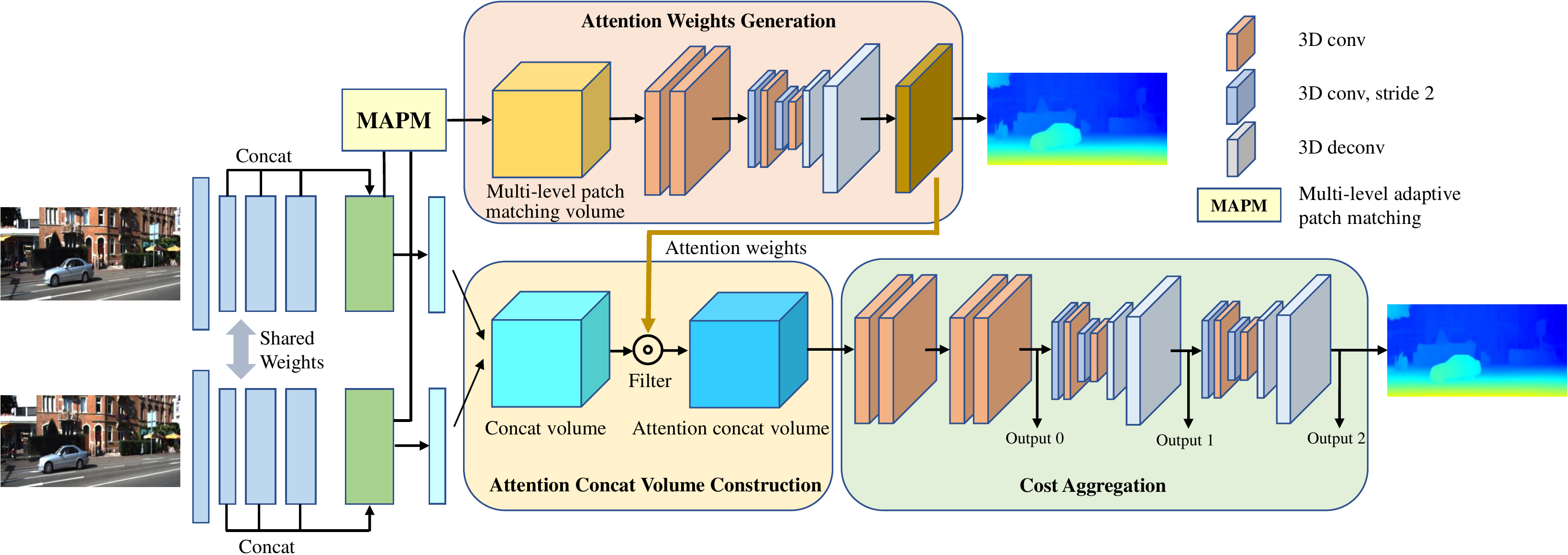} 
\caption{The structure of our proposed ACVNet. The construction process of ACV consists of three steps: initial concatenation volume construction, attention weights generation and attention filtering. Exploiting the generated attention weights to filter the initial concatenation volume can suppress redundant information and enhance matching-related information, deriving attention concatenation volume.
}
\label{fig:acvnet}
\vspace{-10pt}
\end{figure*}

With these intuitions in mind, 
we propose an attention concatenation volume (ACV) which exploits a correlation volume to generate attention weights to filter concatenation volume (see Figure \ref{fig:acvnet}). To have a reliable correlation volume, we propose a novel multi-level adaptive patch matching method to produce more accurate similarity measures, which employs multi-size patches with adaptive weights for matching pixels at different feature levels. The ACV can achieve a higher accuracy and meanwhile significantly alleviate the burden of cost aggregation. Experimental results show that after replacing the combined volume of GwcNet with our ACV, only four 3D convolutions for cost aggregation can achieve better accuracy than GwcNet which employs twenty eight 3D convolutions for cost aggregation. Our ACV is a general cost volume representation that can be seamlessly integrated into various 3D CNN stereo models for performance improvement. Results show that after applying our method, PSMNet and GwcNet  can respectively achieve additional a 42\% and 39\% accuracy improvement.

Based on the advantages of the proposed ACV, we design an accurate stereo matching network ACVNet, which ranks the $2^{nd}$ on the KITTI 2012~\cite{geiger2012we} and KITTI 2015~\cite{menze2015joint} benchmark, the $2^{nd}$ on Scene Flow~\cite{dispNetC2016large}, and the $3^{rd}$ on the ETH3D~\cite{schops2017multi} benchmark among all the published methods (see Figure \ref{fig:ranking}). 
It is noteworthy that our ACVNet is the only method that ranks top 3 concurrently on all four datasets above, demonstrating its good generalization ability to various scenes. Regarding the inference speed, our ACVNet is the fastest among the top 10 methods in the KITTI benchmarks.
Meanwhile, we also design a real-time version of ACVNet, named ACVNet-Fast, which outperforms state-of-the-art real-time methods~\cite{stereonet2018, xu2020aanet, deeppruner2019, yao2021decomposition}.
\vspace{0.1cm}

\section{Related work}\label{sec:Related work}
Recently, CNN-based stereo models~\cite{zbontar2015computing, dispNetC2016large,kendall2017end,guo2019group,nie2019multi, yang2019hierarchical} have achieved impressive performance on most of the standard benchmarks. 
Most of them devote to improving the accuracy and efficiency of cost volume construction and cost aggregation, which are the two key steps of stereo matching. 

\textbf{Cost volume construction}. Existing cost volume representation can be roughly categorized into three types: the correlation volume, the concatenation volume and a combined volume by concatenating the two volumes. 
DispNetC~\cite{dispNetC2016large} utilizes a correlation layer to directly measures the similarities of left and right image features to form a single-channel cost volume for each disparity level. Then, 2D convolution is applied to aggregate contextual information. Such full correlation volume demands low memory and computational complexity, yet the encoded information is too limited (i.e. large content information loss in the channel dimension) to achieve a satisfactory accuracy. GC-Net~\cite{kendall2017end} uses the concatenation volume, which concatenate the left and right CNN features to form a 4D cost volume for all disparities. Such 4D concatenation volume preserves abundant content information from all feature channels and thus outperforms the correlation volume in terms of better accuracy. However, as the concatenation volume does not explicitly encodes similarity measures, it requires a deep stack of 3D convolutions to aggregate costs of all disparities from scratch. 
To overcome the above drawbacks, GwcNet~\cite{guo2019group} proposes the group-wise correlation volume and concatenates it with a compact concatenation volume to form a combined volume, which aims to combine the advantages of two volumes.
However, directly concatenating two types of volumes without considering their respective characteristics yields an inefficient use of the complementary strengths in the two volumes. As a result, deep stacking 3D convolutions in the hourglass architecture are still demanded for cost aggregation in GwcNet~\cite{guo2019group}. 

Following the 4D combined cost volume,
cascade cost volumes~\cite{shen2021cfnet,gu2020cascade,wang2021patchmatchnet} further reduce the memory and computational complexity of cost volume construction by building a cost volume pyramid in a coarse-to-fine manner to progressively narrow down the target disparity range and refine the depth map. However, such coarse-to-fine strategy inevitably involves accumulated errors, i.e. errors in a previous stage can hardly be compensated in the latter stages and in turn yield large errors for some cases. While our ACV only adjusts the weights of different disparities. Thus, although the attention weights are imperfect, the concatenation volume which contains rich context, can help amend errors via the subsequent aggregation network.

\textbf{Cost aggregation}. The goal of this step is to aggregate contextual information in the initial cost volume to derive accurate similarity measures. Many existing methods ~\cite{chang2018pyramid, nie2019multi,chabra2019stereodrnet} exploit a deep 3D CNN to learn an effective similarity function from the cost volume. However, the computational and memory consumption is too high for time-constrained applications. To reduce the complexity, AANet~\cite{xu2020aanet} proposes an intra-scale and cross-scale cost aggregation algorithm to replace the conventional 3D convolutions which can achieve very fast inference speed with a sacrifice of nontrivial accuracy degradation. GANet~\cite{zhang2019ga} also tries to replace 3D convolutions with two guided aggregation layers, which achieves a higher accuracy using spatially dependent 3D aggregation, but at the cost of a higher aggregation time due to the two guided aggregation layers. Even, their final model still uses fifteen 3D convolutions.

Cost volume construction and aggregation are two tightly-coupled modules which jointly determine the accuracy and efficiency of a stereo matching network. In this work, we propose a highly efficient yet informative cost volume representation, named attention concatenation volume, by using the similarity information encoded in the correlation volume to regularize the concatenation volume so that only a lightweight aggregation network is demanded to achieve an overall high efficiency and accuracy. 

\section{Method}\label{sec:method}

\subsection{Attention concatenation volume} \label{sec:acv}
The construction process of attention concatenation volume (ACV) consists of three steps: initial concatenation volume construction, attention weights generation and attention filtering.

\textbf{Initial concatenation volume construction.} Given an input stereo image pair whose  size is $H{\times}W{\times}3$, for each image, we obtain unary feature maps $\mathbf{f}_l$ and $\mathbf{f}_r$ for the left and right images respectively from CNN feature extraction. The size of feature maps of $\mathbf{f}_l$ ($\mathbf{f}_r$) is $N_c{\times}H/4{\times}W/4$ $(N_c=32)$. The initial concatenation volume is then formed by concatenating the $\mathbf{f}_l$ and $\mathbf{f}_r$ for each disparity level as,
\begin{equation}
\mathbf{C}_{concat}(\cdot,d,x,y)=\text{Concat}\left\{\mathbf{f}_{l}(x,y),\mathbf{f}_{r}(x-d,y)\right\},
\end{equation}
the accordingly size of $\mathbf{C}_{concat}$ is $2N_c\times{D}/4\times{H}/4\times{W}/4$, $D$ denotes the maximum of disparity.

\begin{figure}
\centering
{\includegraphics[width=1.0\linewidth]{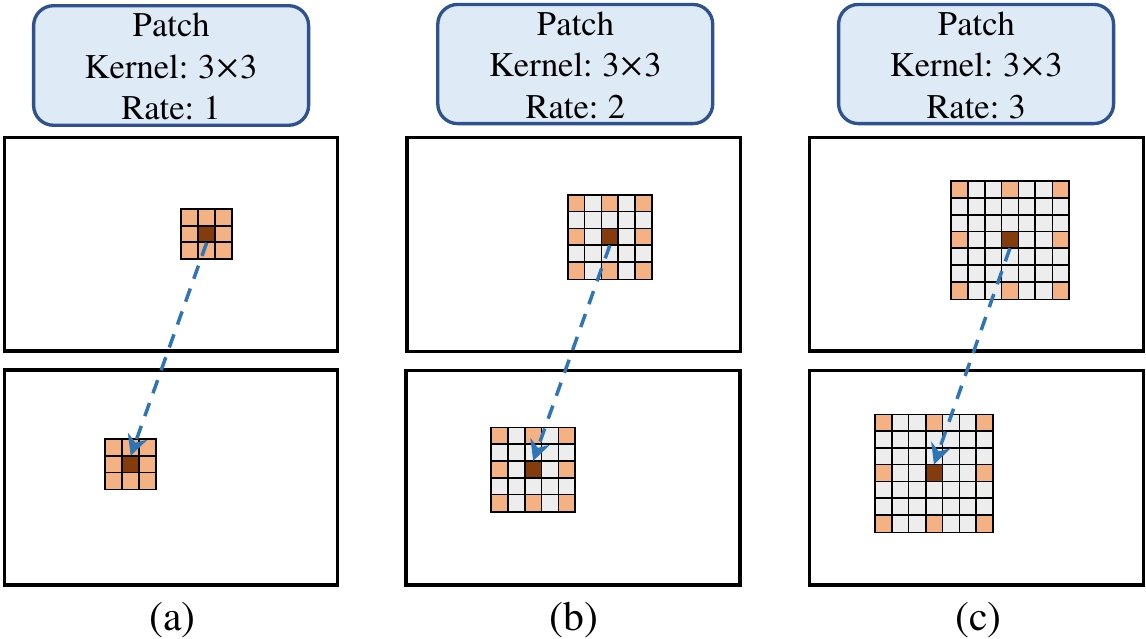}}
\caption{
Multi-level adaptive patch matching. Atrous patches with kernel size 3 $\times$ 3 and different rates can adaptively learn weights for different level. (a), (b) and (c) are for three-level feature maps respectively, $l_1$, $l_2$ and $l_3$. Leveraging large-size patch to include  more  contextual  information  to  better  distinguish the matching costs of different disparities for high-level feature maps.}\label{fig:patch}
\vspace{-10pt}
\end{figure}

\textbf{Attention weights generation.} \label{sec:att_weighs} The attention weights aim to filter the initial concatenation volume so as to emphasize useful information and suppress irrelevant information. To this end, we generate attention weights by extracting geometric information from correlations between a pair of stereo images. Conventional correlation volume is obtained by computing pixel-to-pixel similarity which becomes unreliable for textureless regions due to lack of sufficient matching clues. To address this problem, we propose a more robust correlation volume construction method via multi-level adaptive patch matching (MAPM). Figure \ref{fig:patch} illustrates the key idea of our MAPM. We obtain feature maps at three different levels $l_1$, $l_2$ and $l_3$ from the feature extraction module, and the number of channels for $l_1$, $l_2$ and $l_3$ is 64, 128 and 128 respectively. For each pixel at a particular level, we utilize an atrous patch with a predefined size and adaptively learned weights to calculate the matching cost. By controlling the dilation rate, we ensure that the patch’s scope is related to feature map level and meanwhile maintains the same number of pixels in similarity calculation for the center pixel. The similarity of two corresponding pixels is then a weighted sum of correlations between corresponding pixels within in the patch (denoted by red and orange colors in Figure \ref{fig:patch}).

We adopt the group-wise idea of GwcNet~\cite{guo2019group} to split features into groups and compute correlation maps group by group. Three levels feature maps of $l_1$, $l_2$ and $l_3$ are concatenated to form $N_f$-channel unary feature maps ($N_f$=320). We equally divide $N_f$ channels into $N_g$ groups ($N_g$=40), and accordingly the first 8 groups are from $l_1$, the middle 16 groups are from $l_2$, and the last 16 groups are from $l_3$. Feature maps of different levels will not interfere with each other. We denote the $gth$ feature group as $\mathbf{f}_l^g$, $\mathbf{f}_r^g$, and multi-level patch matching volume $\mathbf{C}_{patch}$ is computed as,
\begin{equation}
\begin{split}
\mathbf{C}_{patch}^{l_{k}}(g,d,x,y)=\frac{1}{N_f/N_g} \sum\limits_{(i,j)\in\Omega^k}\omega_{ij}^{k,g}\cdot C_{ij}^g(d,x,y) \\
C_{ij}^g(d,x,y)=\langle\mathbf{f}_l^g(x\!-\!i,y\!-\!j), \mathbf{f}_{r}^g(x\!-\!i-\!d,y\!-\!j)\rangle,
\end{split}
\end{equation}
where $\mathbf{C}_{patch}^{l_{k}}$ ($k$$\in$$(1,2,3)$) represents matching cost of different feature level $k$. $\langle\cdot,\cdot\rangle$ is the inner product, $(x,y)$ represents the pixel's location, and $d$ denotes different disparity level. $\Omega^k$=$(i,j)$ ($i,j$$\in$$(-k,0,k)$) is a nine-point coordinate set, defining the scope of the patch on the $k$-level feature maps (denoted by red and orange pixels in Figure \ref{fig:patch} ($k$$\in$$(1,2,3)$). $\omega_{ij}^{k}$ represents the weight of a pixel $(i,j)$ in the patch on the $k$-level feature maps and is learned adaptively during the training process. The final multi-level patch matching volume is then obtained by concatenating matching costs $\mathbf{C}_{patch}^{l_k}$ ($k$$\in$$(1,2,3)$ of all levels,
\begin{equation}
\mathbf{C}_{patch}=\text{Concat}\left\{\mathbf{C}_{patch}^{l_1},\mathbf{C}_{patch}^{l_2},\mathbf{C}_{patch}^{l_3}\right\},
\end{equation}
we denote the derived multi-level patch matching volume as $\mathbf{C}_{patch}\in\mathbb{R}^{N_g\times{D}/4\times{H}/4\times{W}/4}$,
we then apply two 3D convolutions and a 3D hourglass network~\cite{guo2019group}
to regularize $\mathbf{C}_{patch}$, and then use another convolution layer to compress the channels to 1 and derive the attention weights, i.e. $\mathbf{A}\in\mathbb{R}^{1\times{D}/4\times{H}/4\times{W}/4}$.

To obtain accurate attention weights of different disparity to filter the initial concatenation volume, we use ground truth disparity to supervise $\mathbf{A}$. Specifically, we 
adopt the same soft argmin function (in Equ. \ref{equ:soft-argmin}) as GC-Net~\cite{kendall2017end} to obtain the disparity estimation $\mathbf{d}_{att}$ from $\mathbf{A}$. We compute smooth L1 loss between $\mathbf{d}_{att}$ and disparity ground truth to guide network learning process for deriving accurate attention weights $\mathbf{A}$.

\textbf{Attention filtering.} After obtaining the attention weights $\mathbf{A}$, we use it to eliminate redundant information in the initial concatenation volume and in turn enhance its representation ability. The attention concatenation volume $\mathbf{C}_{ACV}$ at channel $i$ is computed as,
\begin{equation}
\mathbf{C}_{ACV}(i)=\mathbf{A}\odot\mathbf{C}_{concat}(i),
\label{equ:acv}
\end{equation}
where $\odot$ represents the element-wise product, and the attention weights $\mathbf{A}$ is applied to all channels of the initial concatenation volume.

\subsection{ACVNet architecture} \label{sec:architecture}
Based on the ACV, we design an accurate and efficient end-to-end stereo matching network, named ACVNet. Figure \ref{fig:acvnet} shows the architecture of our ACVNet which consists of four modules of unary feature extraction, attention concatenation volume construction, cost aggregation and disparity prediction. In the following, we introduce each module in details.

\textbf{Feature extraction.} We adopt the three-level ResNet-like architecture in ~\cite{guo2019group}. For the first three layers, three convolutions of $3{\times}3$ kernel with strides of 2, 1 and 1 are used to downsample the input images. Then, 16 residual layers~\cite{he2016deep} are followed to produce unary features at 1/4 resolution, i.e. $l_1$, 6 residual layers with more channels are followed to obtained large receptive fields and semantic information, i.e. $l_2$ and $l_3$. Finally, all feature maps ($l_1$, $l_2$, $l_3$) at 1/4 resolution are concatenated to form 320-channel feature maps for the generation of attention weights. Then two convolutions are applied to compress the 320-channel feature maps to 32-channel feature maps for construction of the initial concatenation volume, which are denoted as $\mathbf{f}_l$ and $\mathbf{f}_r$.

\textbf{Attention concatenation volume construction.} This module takes the 320-channels feature maps for attention weights generation, and $\mathbf{f}_l$ and $\mathbf{f}_r$ for initial concatenation volume construction. Then attention weights are used to filter the initial concatenation volume to produce a 4D cost volume for all disparities, as described in Section \ref{sec:acv}.

\textbf{Cost aggregation.} We process the ACV using a pre-hourglass module which consists of four 3D convolutions with batch normalization and ReLU, and two stacked 3D hourglass networks~\cite{guo2019group}, each of which mainly consists of four 3D convolutions and two 3D deconvolutions stacked in an encoder-decoder architecture, see Figure \ref{fig:acvnet}. 

\textbf{Disparity prediction.} Three outputs are obtained in the cost aggregation, see Figure \ref{fig:acvnet}. For each output, following GwcNet~\cite{guo2019group}, two 3D convolutions are employed to output a 1-channel 4D volume, then we upsample and convert it into a probability volume by softmax function along the disparity dimension. Finally, the predicted value is computed by the soft argmin function~\cite{kendall2017end},
\begin{equation}
d=\sum\limits_{k=0}^{D_{max}-1}k\cdot{p_{k}},
\label{equ:soft-argmin}
\end{equation}
where $k$ denotes disparity level, $p_k$ denotes the corresponding probability. The three predicted disparity maps are denoted as $\mathbf{d}_0$, $\mathbf{d}_1$, $\mathbf{d}_2$.

\begin{figure}
\centering
{\includegraphics[width=1.0\linewidth]{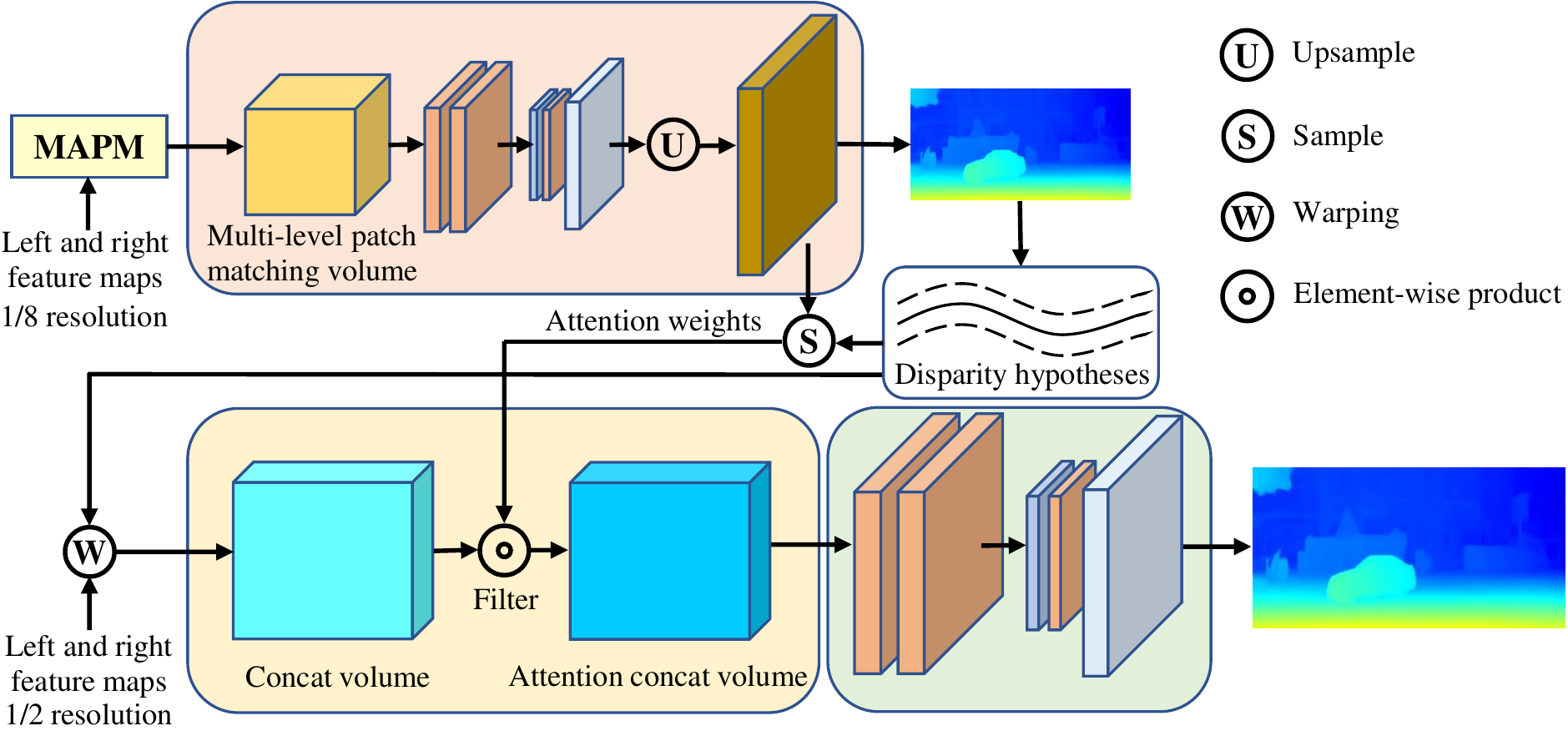}}
\caption{
The structure of ACVNet-Fast.}\label{fig:acv_fast}
\vspace{-10pt}
\end{figure}

\subsection{ACVNet-Fast} \label{sec:acv_fast}
We also construct a real-time version of ACVNet, named ACVNet-Fast. 
ACVNet-Fast adopts the same feature extraction but with fewer layers and disparity prediction modules as ACVNet. Figure \ref{fig:acv_fast} shows the architecture of ACVNet-Fast, the main differences between ACVNet-Fast and ACVNet lie in ACV construction and aggregation. 

Specially, we construct the multi-level patch matching volume based on 1/8 resolution feature maps, and then use two 3D convolutions and a 3D hourglass network to regularize it to obtain the attention weights of 1/8 resolution i.e. $\mathbf{A}^{f}\in\mathbb{R}^{1\times{D}/8\times{H}/8\times{W}/8}$. To achieve real-time performance without sacrificing too much accuracy, we narrow disparity search space by sampling $h$($h$=6) hypotheses near the predicted disparity $\mathbf{d}_{att}^{f}\in\mathbb{R}^{H/2\times{W/2}}$ obtained by upsampled attention weights at 1/2 resolution. These hypotheses $\mathbf{D}_{hyp}\in\mathbb{R}^{h\times{H/2}\times{W/2}}$ is uniformly sampled within the range of $(\mathbf{d}_{att}^{f}-h/2, \mathbf{d}_{att}^{f}+h/2)$. According to these hypotheses, we construct sparse concatenation volume and sample attention weights to get sparse attention weights. Then we construct sparse attention 
concatenation volume $\mathbf{C}_{ACV}^s\in\mathbb{R}^{2N_c^f\times{6}\times{H}/2\times{W}/2}$ by Equ. \ref{equ:acv}.

For cost aggregation, we only use two 3D convolutions and one 3D hourglass network to regularize $\mathbf{C}_{ACV}^s$. As the matching information contained in $\mathbf{C}_{ACV}^s$ is very effective, only a very lightweight aggregation network is required.
In this way, we achieve a good balance of accuracy and speed.

\subsection{Loss function} \label{sec:loss}
For ACVNet, the final loss is given by,
\begin{equation}
\begin{split}
	L=\lambda_{att}\cdot\text{Smooth}_{L_1}(\mathbf{{d}}_{att}-\mathbf{d}^{gt})+\\ \sum_{i=0}^{i=2}\lambda_i\cdot\text{Smooth}_{L_1}(\mathbf{d}_i-\mathbf{d}^{gt}),
\end{split}
\label{equ:loss}
\end{equation}
where $\mathbf{d}_{att}$ is obtained by attention weights $\mathbf{A}$ in Section \ref{sec:acv}. $\lambda_{att}$ represents the coefficient for the predicted $\mathbf{{d}}_{att}$,  $\lambda_i$ represents the coefficient for the $ith$ predicted disparity and $\mathbf{d}^{gt}$ denotes the ground-truth disparity map. 
For ACVNet-Fast, the final loss is given by,
\begin{equation}
	L^{f}=\lambda_{att}^f\cdot\text{Smooth}_{L_1}(\mathbf{{d}}_{att}^f-\mathbf{d}^{gt})+\lambda^f\cdot\text{Smooth}_{L_1}(\mathbf{{d}}^f-\mathbf{d}^{gt}),
\label{equ:loss}
\end{equation}
where $\mathbf{d}^f$ is final output of ACVNet-Fast. The $\text{Smooth}_{L_1}$ is the smooth L1 loss.

\begin{table*} 
\begin{center}
\begin{tabular}{l|c|c|c|c|c|c|c|c|c|c}
\hline
\multirow{2}{*}{Model} & Patch & Multi-level & Attention   & Hourglass & Supervise & \textgreater1px & \textgreater2px &\textgreater3px & D1 & EPE\\ 
 & Match   & Adaptive & Weights & for Att & for Att & (\%) & (\%)& (\%) & (\%) & (px) \\
\hline
GwcNet~\cite{guo2019group} & & & & & &8.03 &4.47 &3.30 & 2.71 & 0.76 \\
Gwc-p &\checkmark & & & & & 7.61 &4.25 &3.14 & 2.55 & 0.72 \\
Gwc-mp & \checkmark& \checkmark & & & &7.03 &3.85 &2.78& 2.31 & 0.64 \\
\hline
Gwc-mp-att &\checkmark & \checkmark & \checkmark & & &6.14 &3.39 &2.49& 2.03 & 0.57 \\
Gwc-mp-att-hg &\checkmark& \checkmark & \checkmark & \checkmark & &5.67 &3.09 &2.23& 1.87 & 0.52 \\
Gwc-mp-att-hg-s &\checkmark& \checkmark & \checkmark & \checkmark & \checkmark &\textbf{4.89} &\textbf{2.69} &\textbf{1.98} & \textbf{1.55} & \textbf{0.46} \\
\hline
\end{tabular}
\end{center}
\vspace{-10pt}
\caption{Ablation study of the ACV on Scene Flow~\cite{dispNetC2016large}. $p$ denotes the ordinary patch which has the same rate (rate=1) and equal weights. $mp$ denotes multi-level adaptive patch (Figure \ref{fig:patch}) which has different rates and adaptive weights for three-level feature maps. }\label{tab:acv}
\vspace{-10pt}
\end{table*}

\section{Experiment} \label{sec:experiment}
In this section, we present ablation studies to explore different designs of the ACV, analyze computational complexity and demonstrate the universality of ACV. Finally we evaluate the proposed models on multiple datasets, such as Scene Flow~\cite{dispNetC2016large}, KITTI~\cite{geiger2012we, menze2015joint} and ETH3D~\cite{schops2017multi}. 
\subsection{Datasets and evaluation metrics} \label{sec:Datasets}

\textbf{Scene Flow} is a collection of synthetic stereo datasets which provides 35454 training image pairs and 4370 testing image pairs with the resolution of 960$\times$540. This dataset provides dense disparity maps as ground truth. For Scene Flow dataset, we utilized the widely-used evaluation metrics end-point error (EPE) and percentage of disparity outliers D1. The outliers are defined as the pixels whose disparity errors are greater than $\text{max}(3\text{px}, 0.05d^{*})$, where $d^{*}$ denotes the ground-truth disparity. 

\textbf{KITTI} includes KITTI 2012~\cite{geiger2012we} and KITTI 2015~\cite{menze2015joint}. KITTI 2012 and 2015 are datasets for real-world driving scenes. KITTI 2012 contains 194 training stereo image pairs and 195 testing images pairs, and KITTI 2015 contains 200 training stereo image pairs and 200 testing image pairs. Both datasets provide sparse ground-truth disparities obtained with LIDAR. The resolution of KITTI 2015 is 1242$\times$375, and that of KITTI 2012 is 1226$\times$370. 


\textbf{ETH3D} is a collection of grayscale stereo pairs from indoor and outdoor scenes. It contains 27 training and 20 testing image pairs with sparse labeled ground-truth. Its disparity range is just in the range of 0-64. The percentage of pixels with errors larger than 2 pixels (bad 2.0) and 1 pixel (bad 1.0) are reported.

\begin{table} 
\begin{center}
\small
\begin{tabular}{c|c|c|c|c|c}
\hline
Model & Acv & \makecell{Hourglass \\ number} & \makecell{D1\\(\%)} & \makecell{EPE\\(px)} & \makecell{Params.\\(M)} \\ 
\hline
GwcNet~\cite{guo2019group} &  & 3 & 2.71 & 0.76 & 6.91 \\
Gwc-acv-3 & \checkmark & 3 & \textbf{1.55} & \textbf{0.46} & 7.40 \\
Gwc-acv-1 & \checkmark & 1 & 1.79 & 0.53 & 5.04 \\
Gwc-acv-0 & \checkmark & 0 & 2.08 & 0.59 & \textbf{3.86} \\
\hline
\makecell{\textbf{ACVNet}\\(Gwc-acv-2)} & \checkmark & 2 & 1.59 & 0.48 & 6.22 \\
\hline
\end{tabular}
\end{center}
\vspace{-10pt}
\caption{Computational complexity and accuracy analysis on Scene Flow~\cite{dispNetC2016large}} \label{tab:acv_complexity}
\vspace{-10pt}
\end{table}

\begin{table} 
\begin{center}
\begin{tabular}{l|c|c}
\hline
Model  & D1 (\%) & EPE (px)\\ 
\hline
PSMNet~\cite{chang2018pyramid} & 3.89 & 1.09 \\
PSMNet-ACV & \textbf{2.17} & \textbf{0.63} \\
\hline
GwcNet~\cite{guo2019group} & 2.71 & 0.76 \\
GwcNet-ACV & \textbf{1.55} & \textbf{0.46} \\
\hline
CFNet~\cite{shen2021cfnet} & 4.51 & 0.97 \\
CFNet-ACV & \textbf{4.02} & \textbf{0.83} \\
\hline
\end{tabular}
\end{center}
\vspace{-10pt}
\caption{Universality study of ACV on Scene Flow~\cite{dispNetC2016large}.}\label{tab:universality}
\vspace{-10pt}
\end{table}

\subsection{Implementation details} \label{sec:Implementation details}
We implement our methods with PyTorch and perform our experiment using NVIDIA RTX 3090 GPUs. For all the experiments, we use the Adam~\cite{kingma2014adam} optimizer, with $\beta_1=0.9$, $\beta_2=0.999$. For ACVNet, the coefficients of four outputs are set as $\lambda_{att}$=0.5, $\lambda_{0}$=0.5, $\lambda_{1}$=0.7, $\lambda_{2}$=1.0. For ACVNet-Fast, the coefficients of two outputs are set as $\lambda_{att}^f$=0.5, $\lambda^{f}$=1.0. For Scene Flow, we first train attention weights generation network for 64 epochs and then train the remaining network for another 64 epochs. Finally we train complete network for 64 epochs. The initial learning rate is set to 0.001 decayed by a factor of 2 after epoch 20, 32, 40, 48 and 56. For KITTI, we finetune the pre-trained Scene Flow model on the mixed KITTI 2012 and KITTI 2015 training sets for 500 epochs. Then another 500 epochs are trained on the separate KITTI 2012/2015 training set. The initial learning rate is 0.001 and decreases by half at the 300th epoch.


\subsection{Ablation study} \label{sec:Ablation study}

\begin{figure*}
\centering
\includegraphics[width=1.0\textwidth]{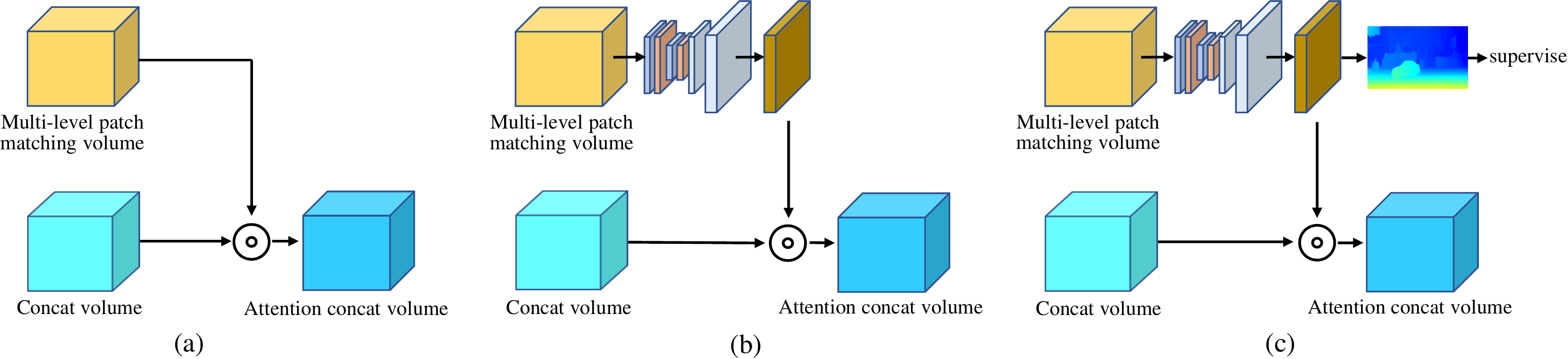} 
\caption{Illustration of different ways of constructing attention concatenation volume (ACV).}
\label{fig:three_acv}
\vspace{-10pt}
\end{figure*} 

\begin{table} 
\begin{center}
\small
\begin{tabular}{l|ccc|ccc}
\hline
\multirow{3}{*}{Model} & \multicolumn{3}{c|}{Scene Flow} & \multicolumn{2}{c}{KITTI 2015}\\ 
 & EPE &D1 &Params. &D1-bg & D1-all\\ 
&(px) &(\%) &(M) &(\%) & (\%)\\
 
\hline
GwcNet~\cite{guo2019group} & 0.76 &2.71 &6.91 & 1.74 & 2.11\\
Gwc-CAS~\cite{gu2020cascade} & 0.62 & 2.55 &10.77 & 1.59 & 2.00 \\
ACVNet & \textbf{0.48} &\textbf{1.59} &\textbf{6.22} & \textbf{1.37} &\textbf{1.65}\\
\hline
\end{tabular}
\end{center}
\vspace{-10pt}
\caption{Comparisons of ACV and cascaded volume approaches.}\label{tab:acv_cascade}
\vspace{-15pt}
\end{table}

\textbf{Multi-level adaptive patch matching.}  The proposed multi-level adaptive patch matching is a general method that can be applied to most existing stereo models based on the correlation volume. In this study, we take GwcNet~\cite{guo2019group} as a baseline and replace the original point matching based correlation construction method with ordinary patch matching and our multi-level adaptive patch matching to derive three comparison methods, i.e., GwcNet~\cite{guo2019group}, GwcNet-p and GwcNet-mp in Table \ref{tab:acv}. Ordinary patch matching utilizes a fixed size patch (3$\times$3) and equal weights for all pixels in the patch. Results show that only a slight improvement can be achieved by GwcNet-p compared with original GwcNet~\cite{guo2019group}, but the proposed multi-level patch matching has a very significant improvement.

\textbf{Attention concatenation volume.} We evaluate different strategies for constructing the ACV on Scene Flow~\cite{dispNetC2016large}. We still take GwcNet~\cite{guo2019group} as our baseline, replace its combined volume with our ACV and keep the subsequent aggregation and disparity prediction modules the same. Figure \ref{fig:three_acv} shows three different ways of constructing the ACV. Figure \ref{fig:three_acv} (a) directly averages the multi-level patch matching volume along the channel dimension and multiply it with the concatenation volume, denoted as GwcNet-mp-att. As shown in Table \ref{tab:acv}, just this simple approach can dramatically improve the accuracy.
Apparently, when using multi-level patch matching volume to filter concatenation volume, the accuracy of multi-level patch matching volume is crucial and largely affects the final performance of the network, so we use an hourglass architecture of 3D convolutions to aggregate it, which is denoted as GwcNet-mp-att-hg shown in Figure \ref{fig:three_acv} (b). The results in Table \ref{tab:acv} show that GwcNet-mp-att-hg improve the D1 and EPE by 7.9\% and 8.7\% respectively compared with GwcNet-mp-att. To further explicitly constrain multi-level patch matching volume during training, we use the softmax and soft argmin function for regression to obtain the predicted disparity, and use the ground truth to supervise the disparity, denoted as GwcNet-mp-att-hg-s shown in Figure \ref{fig:three_acv} (c). Compared with the GwcNet-mp-att-hg, GwcNet-mp-att-hg-s improves the D1 and EPE by 17.1\% and 11.5\% respectively with no computational cost increase in the inference stage. Overall, by replacing the combined volume in GwcNet~\cite{guo2019group} with our ACV, our GwcNet-mp-att-hg-s model achieves 42.8\% and 39.5\% improvement for D1 and EPE compared with GwcNet~\cite{guo2019group}, demonstrating the effectiveness of ACV.

\subsection{Computational complexity analysis} \label{sec:High efficiency}
An ideal cost volume should require few parameters for subsequent aggregation network and meanwhile enable a satisfactory disparity prediction accuracy. In this subsection, we analyze the complexity of ACV in terms of the number of parameters demanded in the subsequent aggregation network and the corresponding accuracy. We use GwcNet~\cite{guo2019group} as the baseline. In original GwcNet~\cite{guo2019group}, it uses a three stacked hourglass networks for cost aggregation. We first replace the combined volume in the original GwcNet~\cite{guo2019group} with our ACV with other parts remain the same. The corresponding model is denoted as Gwc-acv-3 in Table \ref{tab:acv_complexity}. The results show that compared with GwcNet~\cite{guo2019group}, Gwc-acv-3 improves D1 and EPE by 42.8\% and 39.5\% respectively. We further reduce the number of hourglass networks from 3 to 2, 1, and 0, the correspondingly derived models are denoted as Gwc-acv-2, Gwc-acv-1 and Gwc-acv-0. The results in Table \ref{tab:acv_complexity} show that, as the number of parameters reduced in the aggregation network, the prediction errors slightly increase. 
More importantly, after using our ACV, the stereo model without any hourglass network, i.e., Gwc-acv-0, even outperforms GwcNet.
To achieve a both high accuracy and efficiency, we choose Gwc-acv-2 as our final model, and we denote it as ACVNet.

\subsection{Universality and superiority of ACV} \label{sec:Universality}
To demonstrate the universality of our ACV, we integrate our ACV into three state-of-the-art models, i.e. GwcNet~\cite{guo2019group}, PSMNet~\cite{chang2018pyramid} and CFNet~\cite{shen2021cfnet}, and compare the performance of the original models with those after using our ACV. Specifically, we denote the model after applying our method as GwcNet-ACV, PSMNet-ACV and CFNet-ACV for comparison respectively. As shown in Table \ref{tab:universality}, the EPE is reduced by 39.5\% for GwcNet~\cite{guo2019group}, 42.2\% for PSMNet~\cite{chang2018pyramid} and 14.4\% for CFNet~\cite{shen2021cfnet}.

We experimentally compare our ACV with cascaded approaches. We apply the two-stage cascaded method proposed by ~\cite{gu2020cascade} to GwcNet, the corresponding model is denoted as Gwc-CAS. As shown in Table \ref{tab:acv_cascade}, our ACV outperforms cascaded approach. We think
the superior performance of ACV to the cascaded approach
is because the latter could suffer from irreversible cumulative errors as it directly discards disparities that is beyond
the prediction range. However, our ACV only adjusts the
weights of different disparities. Thus, although the attention weights are imperfect, the concatenation volume which
contains rich context, can help amend errors to some extend
via the subsequent aggregation network.
\begin{table} 
\begin{center}
\small
\begin{tabular}{c|c|cc}
\hline
\multirow{2}{*}{Model} & Scene Flow & \multicolumn{2}{c}{ETH3D}\\ 
 & EPE (px) & bad 1.0 (\%) & bad 2.0 (\%)\\ 
\hline
PSMNet~\cite{chang2018pyramid} & 1.09& 5.02 & 1.09 \\
GANet~\cite{zhang2019ga} & 0.84 & 6.56 & 1.10 \\
CFNet~\cite{shen2021cfnet} & 0.97 & 3.31 & \underbar{0.77} \\
LEAStereo~\cite{cheng2020hierarchical} & 0.78 & - & - \\
HITNet~\cite{tankovich2021hitnet} & \textbf{0.43} & \underbar{2.79} & 0.80 \\
ACVNet (ours) & \underbar{0.48} & \textbf{2.58} & \textbf{0.57} \\
\hline
\end{tabular}
\end{center}
\vspace{-10pt}
\caption{Quantitative evaluation on Scene Flow~\cite{dispNetC2016large} and ETH3D~\cite{schops2017multi}. \textbf{Bold}: Best, \underbar{Underscore}: Second best.}\label{tab:acv_scene_eth}
\end{table}

\begin{table} 
\begin{center}
\setlength{\tabcolsep}{2pt}
\small
\begin{tabular}{c|c|c|c|c}
\hline
\multirow{2}{*}{Model} & Scene Flow & KITTI 12 & KITTI 15 & Time\\ 
  &EPE (px) & 3-Noc (\%) & D1-all (\%) & (ms) \\ 
\hline
StereoNet~\cite{stereonet2018} & 1.10 & - & 4.83 & \textbf{15} \\
DeepPrunerFast~\cite{deeppruner2019} & 0.97 & - & 2.59 & 61 \\
AANet~\cite{xu2020aanet} & 0.87 & 1.91 & 2.55 & 62 \\
DecNet~\cite{yao2021decomposition} & \underbar{0.84} & - & 2.37 & 50 \\
HITNet~\cite{tankovich2021hitnet} & -  & \textbf{1.41} & \textbf{1.98} & \underbar{20}\\
ACVNet-Fast (ours) & \textbf{0.77} & \underbar{1.82} & \underbar{2.34} & 48 \\
\hline
\end{tabular}
\end{center}
\vspace{-10pt}
\caption{ACVNet-Fast performance on Scene Flow and KITTI.}
\vspace{-10pt}
\label{tab:acv_fast}
\end{table}

\begin{table*}
    \centering
    \begin{tabular}{c|cccccc|ccc|c}
    \hline
     & \multicolumn{6}{c|}{KITTI 2012~\cite{geiger2012we}} & \multicolumn{3}{c|}{ KITTI 2015~\cite{menze2015joint}} & \\
    \hline
    Method& 2-noc & 2-all & 3-noc & 3-all & \thead{EPE \\ noc} & \thead{EPE\\all} & D1-bg & D1-fg & D1-all & \makecell{Run-time\\(s)} \\
    \hline
    {GC-Net~\cite{kendall2017end}}  & {2.71} & {3.46} &1.77 & 2.30 & 0.6 & 0.7 & 2.21 & 6.16 & 2.87 & 0.9 \\
    {PSMNet~\cite{chang2018pyramid}}  & {2.44} & {3.01} &1.49 & 1.89 & 0.5 & 0.6 & 1.86 & 4.62 & 2.32 & 0.41 \\
    EdgeStereo~\cite{song2018edgestereo}  & 2.32 & 2.88 & 1.46 & 1.83 & \textbf{0.4} & \textbf{0.5}  & 1.84 & 3.30 & 2.08 & 0.32 \\
    GwcNet~\cite{guo2019group} & 2.16 & 2.71 & 1.32 & 1.70 & 0.5 & \textbf{0.5}  & 1.74 & 3.93 & 2.11 & 0.32 \\
    GANet-deep~\cite{zhang2019ga}  & 1.89 & 2.50 & 1.19 & 1.60 & \textbf{0.4} &\textbf{0.5} & 1.48 & 3.46 & 1.81 & 1.8 \\
    AcfNet~\cite{zhang2020adaptive} & \textbf{1.83} & \textbf{2.35} &{1.17}  & {1.54} &  {0.5} & \textbf{{0.5}} & {1.51}  & {3.80} & {1.89} & 0.48\\
    HITNet~\cite{tankovich2021hitnet} & {2.00} & {2.65} &{1.41}  & {1.89} &  \textbf{{0.4}} & \textbf{{0.5}} & {1.74}  & {3.20} & {1.98} & 0.02\\
    CFNet~\cite{shen2021cfnet} & {1.90} & {2.43} &{1.23}  & {1.58} &\textbf{0.4} & \textbf{{0.5}} & {1.54}  & {3.56} & {1.88} & 0.18\\
    LEAStereo~\cite{cheng2020hierarchical}  & 1.90 & 2.39 & \textbf{1.13} & \textbf{1.45} & 0.5 &\textbf{0.5} & \underbar{1.40} & \textbf{2.91} & \textbf{1.65} & 0.3 \\
    ACVNet (ours)  & \textbf{1.83} &  \textbf{2.35} & \textbf{1.13} & \underbar{1.47} & \textbf{0.4} &\textbf{0.5} & \textbf{1.37} & \underbar{3.07} & \textbf{1.65} & 0.2 \\
    \hline
    \end{tabular}
    \caption{Quantitative evaluation on KITTI 2012~\cite{geiger2012we} and KITTI 2015~\cite{menze2015joint}. 
    For KITTI 2012, we report the percentage of pixels with errors larger than $x$ disparities in both non-occluded (x-noc) and all regions (x-all), as well as the overall EPE in both non occluded (EPE-noc) and all the pixels (EPE-all). For KITTI 2015, we report D1 metric in background regions (bg), foreground areas (fg), and all. 
    \textbf{Bold}: Best, \underbar{Underscore}: Second best.}
\label{tab:kitti1215}
\end{table*}

\begin{figure*}
\centering
\includegraphics[width=1.0\textwidth]{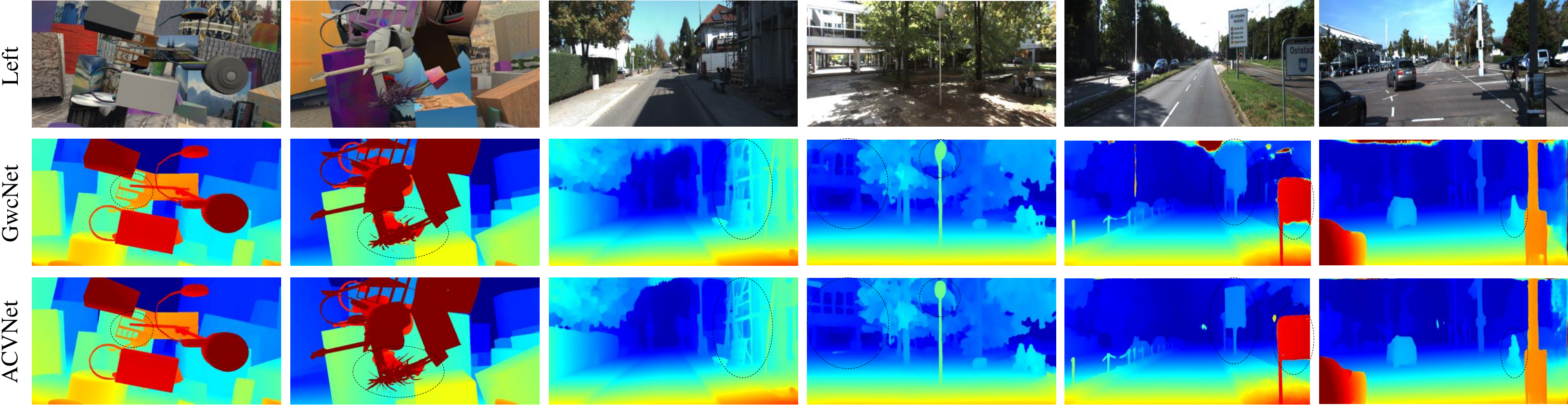} 
\caption{Qualitative results on Scene Flow~\cite{dispNetC2016large} and KITTI~\cite{geiger2012we, menze2015joint}.The first two columns show results on Scene Flow, the middle two columns show results on KITTI 2012, and the last two columns show results on KITTI 2015.}
\label{fig:sceneflow_kt}
\vspace{-15pt}
\end{figure*}

\subsection{ACVNet performance} \label{sec:acv_benchmark}
\textbf{Scene Flow.} 
As shown in Table \ref{tab:acv_scene_eth}, our method achieves state-of-the-art performance. We can observe that our ACVNet improves EPE accuracy by 38.4\% and meanwhile has a faster inference speed compared with the state-of-the-art method LEAStereo\cite{cheng2020hierarchical}, i.e. 0.2s vs. 0.3s.

\textbf{KITTI.} As shown in Table \ref{tab:kitti1215} and Figure \ref{fig:ranking}, our ACVNet outperforms most  existing  published methods and ranks No.2 in KITTI 2012 and KITTI 2015 leaderboards. 
It is worth mentioning that our ACVNet is also the fastest among the top 10 methods in the KITTI benchmark leaderboards.

\textbf{ETH3D.} As shown in Table \ref{tab:acv_scene_eth}, our ACVNet outperforms the state-of-the-art methods, HITNet~\cite{tankovich2021hitnet} and CFNet~\cite{shen2021cfnet}.

To sum up, our ACVNet shows excellent performance on the above four datasets, and it is worth mentioning that our ACVNet is also the only method that ranks top 5 concurrently in all four datasets, which represents the good generalization ability of our method to various scenes. The current SOTA methods always perform poorly in some certain scenarios, e.g. LEAStereo\cite{cheng2020hierarchical} has poor accuracy on Scene Flow; the performance of HITNet~\cite{tankovich2021hitnet} in real-world scenarios (KITTI and ETH3D) is far inferior to our ACVNet.

\subsection{ACVNet-Fast performance} \label{sec:AcvNet-fast performance}
To demonstrate the outstanding performance of our ACVNet-Fast, we compared it with the current classic real-time networks on Scene Flow~\cite{dispNetC2016large} and KITTI~\cite{geiger2012we, menze2015joint} benchmark. As shown in Table \ref{tab:acv_fast}, our method achieves a very good balance between inference time and accuracy.

\section{Conclusion} \label{sec:conclusion}
In this paper, we propose a novel cost volume, named attention concatenation volume (ACV), which generates attention weights based on similarity measures to filter concatenation volume. We also propose a novel multi-level adaptive patch matching method to produce accurate similarity measures even for textureless regions. Based on ACV, we design a highly accurate network (ACVNet), which 
shows excellent performance
on four public benchmarks, i.e., KITTI 2012\&2015, Scene Flow and ETH3D. 

\textbf{Acknowledgements.} This work is supported by National Natural Science Foundation of China (62122029), WNLOK Open Project(2018WNLOKF025).

{\small
\bibliographystyle{ieee_fullname}
\bibliography{egbib}

\begin{thebibliography}{10}\itemsep=-1pt

\bibitem{chabra2019stereodrnet}
Rohan Chabra, Julian Straub, Christopher Sweeney, Richard Newcombe, and Henry
  Fuchs.
\newblock Stereodrnet: Dilated residual stereonet.
\newblock In {\em Proceedings of the IEEE/CVF Conference on Computer Vision and
  Pattern Recognition}, pages 11786--11795, 2019.

\bibitem{chang2018pyramid}
Jia-Ren Chang and Yong-Sheng Chen.
\newblock Pyramid stereo matching network.
\newblock In {\em Proceedings of the IEEE Conference on Computer Vision and
  Pattern Recognition}, pages 5410--5418, 2018.

\bibitem{cheng2020hierarchical}
Xuelian Cheng, Yiran Zhong, Mehrtash Harandi, Yuchao Dai, Xiaojun Chang, Tom
  Drummond, Hongdong Li, and Zongyuan Ge.
\newblock Hierarchical neural architecture search for deep stereo matching.
\newblock {\em arXiv preprint arXiv:2010.13501}, 2020.

\bibitem{deeppruner2019}
Shivam Duggal, Shenlong Wang, Wei-Chiu Ma, Rui Hu, and Raquel Urtasun.
\newblock Deeppruner: Learning efficient stereo matching via differentiable
  patchmatch.
\newblock In {\em Proceedings of the IEEE International Conference on Computer
  Vision}, pages 4384--4393, 2019.

\bibitem{geiger2012we}
Andreas Geiger, Philip Lenz, and Raquel Urtasun.
\newblock Are we ready for autonomous driving? the kitti vision benchmark
  suite.
\newblock In {\em 2012 IEEE conference on computer vision and pattern
  recognition}, pages 3354--3361. IEEE, 2012.

\bibitem{gu2020cascade}
Xiaodong Gu, Zhiwen Fan, Siyu Zhu, Zuozhuo Dai, Feitong Tan, and Ping Tan.
\newblock Cascade cost volume for high-resolution multi-view stereo and stereo
  matching.
\newblock In {\em Proceedings of the IEEE/CVF Conference on Computer Vision and
  Pattern Recognition}, pages 2495--2504, 2020.

\bibitem{guo2019group}
Xiaoyang Guo, Kai Yang, Wukui Yang, Xiaogang Wang, and Hongsheng Li.
\newblock Group-wise correlation stereo network.
\newblock In {\em Proceedings of the IEEE/CVF Conference on Computer Vision and
  Pattern Recognition}, pages 3273--3282, 2019.

\bibitem{he2016deep}
Kaiming He, Xiangyu Zhang, Shaoqing Ren, and Jian Sun.
\newblock Deep residual learning for image recognition.
\newblock In {\em Proceedings of the IEEE/CVF Conference on Computer Vision and
  Pattern Recognition}, pages 770--778, 2016.

\bibitem{kendall2017end}
Alex Kendall, Hayk Martirosyan, Saumitro Dasgupta, Peter Henry, Ryan Kennedy,
  Abraham Bachrach, and Adam Bry.
\newblock End-to-end learning of geometry and context for deep stereo
  regression.
\newblock In {\em Proceedings of the IEEE International Conference on Computer
  Vision}, pages 66--75, 2017.

\bibitem{stereonet2018}
Sameh Khamis, Sean Fanello, Christoph Rhemann, Adarsh Kowdle, Julien Valentin,
  and Shahram Izadi.
\newblock Stereonet: Guided hierarchical refinement for edge-aware depth
  prediction.
\newblock In {\em European Conference on Computer Vision}, 2018.

\bibitem{kingma2014adam}
Diederik~P Kingma and Jimmy Ba.
\newblock Adam: A method for stochastic optimization.
\newblock {\em arXiv preprint arXiv:1412.6980}, 2014.

\bibitem{dispNetC2016large}
Nikolaus Mayer, Eddy Ilg, Philip Hausser, Philipp Fischer, Daniel Cremers,
  Alexey Dosovitskiy, and Thomas Brox.
\newblock A large dataset to train convolutional networks for disparity,
  optical flow, and scene flow estimation.
\newblock In {\em Proceedings of the IEEE Conference on Computer Vision and
  Pattern Recognition}, pages 4040--4048, 2016.

\bibitem{menze2015joint}
Moritz Menze, Christian Heipke, and Andreas Geiger.
\newblock Joint 3d estimation of vehicles and scene flow.
\newblock {\em ISPRS annals of the photogrammetry, remote sensing and spatial
  information sciences}, 2:427, 2015.

\bibitem{nie2019multi}
Guang-Yu Nie, Ming-Ming Cheng, Yun Liu, Zhengfa Liang, Deng-Ping Fan, Yue Liu,
  and Yongtian Wang.
\newblock Multi-level context ultra-aggregation for stereo matching.
\newblock In {\em Proceedings of the IEEE/CVF Conference on Computer Vision and
  Pattern Recognition}, pages 3283--3291, 2019.

\bibitem{schops2017multi}
Thomas Schops, Johannes~L Schonberger, Silvano Galliani, Torsten Sattler,
  Konrad Schindler, Marc Pollefeys, and Andreas Geiger.
\newblock A multi-view stereo benchmark with high-resolution images and
  multi-camera videos.
\newblock In {\em Proceedings of the IEEE Conference on Computer Vision and
  Pattern Recognition}, pages 3260--3269, 2017.

\bibitem{shen2021cfnet}
Zhelun Shen, Yuchao Dai, and Zhibo Rao.
\newblock Cfnet: Cascade and fused cost volume for robust stereo matching.
\newblock In {\em Proceedings of the IEEE/CVF Conference on Computer Vision and
  Pattern Recognition}, pages 13906--13915, 2021.

\bibitem{song2018edgestereo}
Xiao Song, Xu Zhao, Hanwen Hu, and Liangji Fang.
\newblock Edgestereo: A context integrated residual pyramid network for stereo
  matching.
\newblock In {\em Asian Conference on Computer Vision}, pages 20--35. Springer,
  2018.

\bibitem{tankovich2021hitnet}
Vladimir Tankovich, Christian Hane, Yinda Zhang, Adarsh Kowdle, Sean Fanello,
  and Sofien Bouaziz.
\newblock Hitnet: Hierarchical iterative tile refinement network for real-time
  stereo matching.
\newblock In {\em Proceedings of the IEEE/CVF Conference on Computer Vision and
  Pattern Recognition}, pages 14362--14372, 2021.

\bibitem{wang2021patchmatchnet}
Fangjinhua Wang, Silvano Galliani, Christoph Vogel, Pablo Speciale, and Marc
  Pollefeys.
\newblock Patchmatchnet: Learned multi-view patchmatch stereo.
\newblock In {\em Proceedings of the IEEE/CVF Conference on Computer Vision and
  Pattern Recognition}, pages 14194--14203, 2021.

\bibitem{xu2020aanet}
Haofei Xu and Juyong Zhang.
\newblock Aanet: Adaptive aggregation network for efficient stereo matching.
\newblock In {\em Proceedings of the IEEE/CVF Conference on Computer Vision and
  Pattern Recognition}, pages 1959--1968, 2020.

\bibitem{yang2019hierarchical}
Gengshan Yang, Joshua Manela, Michael Happold, and Deva Ramanan.
\newblock Hierarchical deep stereo matching on high-resolution images.
\newblock In {\em Proceedings of the IEEE/CVF Conference on Computer Vision and
  Pattern Recognition}, pages 5515--5524, 2019.

\bibitem{yao2021decomposition}
Chengtang Yao, Yunde Jia, Huijun Di, Pengxiang Li, and Yuwei Wu.
\newblock A decomposition model for stereo matching.
\newblock In {\em Proceedings of the IEEE/CVF Conference on Computer Vision and
  Pattern Recognition}, pages 6091--6100, 2021.

\bibitem{zbontar2015computing}
Jure Zbontar and Yann LeCun.
\newblock Computing the stereo matching cost with a convolutional neural
  network.
\newblock In {\em Proceedings of the IEEE/CVF Conference on Computer Vision and
  Pattern Recognition}, pages 1592--1599, 2015.

\bibitem{zhang2019ga}
Feihu Zhang, Victor Prisacariu, Ruigang Yang, and Philip~HS Torr.
\newblock Ga-net: Guided aggregation net for end-to-end stereo matching.
\newblock In {\em Proceedings of the IEEE/CVF Conference on Computer Vision and
  Pattern Recognition}, pages 185--194, 2019.

\bibitem{zhang2020adaptive}
Youmin Zhang, Yimin Chen, Xiao Bai, Suihanjin Yu, Kun Yu, Zhiwei Li, and
  Kuiyuan Yang.
\newblock Adaptive unimodal cost volume filtering for deep stereo matching.
\newblock In {\em Proceedings of the AAAI Conference on Artificial
  Intelligence}, volume~34, pages 12926--12934, 2020.

\end{thebibliography}
}

\end{document}